\documentclass[a4paper,fleqn]{cas-dc}

\usepackage{threeparttable}
\usepackage[authoryear]{natbib}

\def\tsc#1{\csdef{#1}{\textsc{\lowercase{#1}}\xspace}}
\tsc{WGM}
\tsc{QE}
\tsc{EP}
\tsc{PMS}
\tsc{BEC}
\tsc{DE}

\usepackage{url}
\usepackage{hyperref} 
\usepackage{cleveref} 

\hypersetup{colorlinks = true, 
	    linkcolor = black, 
	    urlcolor = blue,
            citecolor = blue} 

\usepackage{flushend}
\begin{document}
\let\WriteBookmarks\relax
\def\floatpagepagefraction{1}
\def\textpagefraction{.001}

\shorttitle{DeepAAT: Deep Automated Aerial Triangulation for Fast UAV-based Mapping}

\shortauthors{Zequan Chen et~al.}

\title [mode = title]{DeepAAT: Deep Automated Aerial Triangulation for Fast UAV-based Mapping}                      

\author[1]{Zequan Chen}[]
\ead{zeeqchen@whu.edu.cn}

\author[1]{Jianping Li}[orcid=0000-0003-4813-5126]
\cormark[1]
\ead{lijianping@whu.edu.cn}

\author[2]{Qusheng Li}[]
\ead{13878135152@163.com}

\author%
[1]
{Zhen Dong}
\ead{dongzhenwhu@whu.edu.cn}

\author
[1]
{Bisheng Yang}
\ead{bshyang@whu.edu.cn}

\affiliation[1]{organization={LIESMARS, Wuhan University},
    city={Wuhan},
    postcode={430079}, 
    country={China}}

\affiliation[2]{organization={Guangxi Zhuang Autonomous Region Remote Sensing Institute of Nature Resources},
    city={Nanning},
    postcode={530201}, 
    country={China}}

\cortext[cor1]{Corresponding author}

\begin{abstract}
Automated Aerial Triangulation (AAT), aiming to restore image poses and reconstruct sparse points simultaneously, plays a pivotal role in earth observation. AAT has evolved into a fundamental process widely applied in large-scale Unmanned Aerial Vehicle (UAV) based mapping. However classic AAT methods still face challenges like low efficiency and limited robustness. This paper introduces DeepAAT, a deep learning network designed specifically for AAT of UAV imagery. DeepAAT considers both spatial and spectral characteristics of imagery, enhancing its capability to resolve erroneous matching pairs and accurately predict image poses. DeepAAT marks a significant leap in AAT's efficiency, ensuring thorough scene coverage and precision. Its processing speed outpaces incremental AAT methods by hundreds of times and global AAT methods by tens of times while maintaining a comparable level of reconstruction accuracy. Additionally, DeepAAT's scene clustering and merging strategy facilitate rapid localization and pose determination for large-scale UAV images, even under constrained computing resources. The experimental results demonstrate that DeepAAT substantially improves over conventional AAT methods, highlighting its potential for increased efficiency and accuracy in UAV-based 3D reconstruction tasks. To benefit the photogrammetry society, the code of DeepAAT will be released at: \url{https://github.com/WHU-USI3DV/DeepAAT}.
\end{abstract}

\begin{highlights}
\item Incorporating a spatial-spectral feature aggregation module, boosts the network's ability to perceive the spatial distribution of cameras and enhances the global regression capability for camera poses. 
\item Introducing an outlier rejection module according to global consistency, which effectively generates a reliability evaluation score for each feature correspondence.
\item DeepAAT can efficiently process hundreds of UAV images simultaneously, marking a significant breakthrough in enhancing the applicability of deep learning-based AAT algorithms. 
\end{highlights}


\maketitle

\section{Introduction}\label{section_intro}
Automated Aerial Triangulation (AAT) is a basic task in photogrammetry and holds substantial research significance \citep{tanathong2014using}. It serves as the initial step in the 3D reconstruction pipeline of aerial images \citep{ZHONG202316}. AAT's primary role involves simultaneously recovering the camera poses and reconstructing sparse 3D points in the scene. These foundational outputs facilitate subsequent dense image matching and 3D modeling procedures \cite{jiang2021unmanned}. The derived camera poses and scene models find diverse applications in digital mapping \cite{hasheminasab2022linear}, virtual reality \cite{jiang2020efficient}, and smart city \cite{zhou2020selection}. With a research history spanning decades \citep{schenk1997towards}, classical AAT algorithms can be primarily categorized into two groups: incremental style and global style \cite{schonberger2016structure}. Furthermore, the evolution of deep learning has given rise to numerous supervised AAT algorithms \cite{xiao2022deepmle}. The existing AAT methods are reviewed as follows.

\subsection{Classic Automated Aerial Triangulation}
The first step of the classic AAT is to perform feature extraction and matching of all input images. The following steps are different for the global style and incremental style. Global AAT can predict all camera poses and scene structure at once \citep{govindu2004lie}. In AAT algorithms, Bundle Adjustment (BA) \citep{triggs2000bundle} is the most time-consuming part. Global AAT only requires to execute BA once, resulting in higher efficiency. However, it can be difficult to eliminate outliers, resulting in poor robustness and scene integrity. Incremental AAT was first proposed by \citet{snavely2006photo}, with the key lying in selecting a good initial matching image pair \citep{beder2006determining}. Afterward, incremental AAT adds a new image to the system sequentially, followed by Perspective-n-Points (PnP) \citep{lepetit2009ep}, Triangulation \citep{hartley1997triangulation}, and local BA. Incremental AAT requires multiple BA, resulting in low reconstruction efficiency in situations with a large number of images \citep{zhu2017accurate}. In addition, due to the accumulation of errors, the reconstructed scene is prone to drift issues. 

Compared to the general scenes, UAV images exhibit distinctive characteristics, including large volumes, high resolutions, and significant overlap. Within the realm of classic AAT algorithms, incremental methods have emerged as the standard approach for UAV image AAT due to their superior robustness against outliers and ability to provide comprehensive results. Addressing the challenges posed by large-scale UAV images, 
most SOTA AAT methods typically involve employing a divide-and-conquer strategy. This strategy begins by segmenting the UAV image into blocks based on GPS information, followed by the fusion of all modules to yield globally consistent large-scale results. Noteworthy contributions in this field include the work by \citet{chen2020graph}, which employed the maximum spanning tree to expand images after dividing the scene map into smaller segments with a certain degree of overlap, thereby enhancing connectivity and scene map integrity. Similarly, \citet{xu2021robust} introduced a hierarchical approach that constructed a binary tree using images as leaf nodes, subsequently fusing spatial triads and scenes from the bottom up. This method offers advantages in terms of robustness, accuracy, and efficiency. Likewise, \citet{bhowmick2017divide} initially organized images into hierarchical trees using clustering methods, then addressed the AAT problem for large-scale images by reconstructing each small image set and merging them into a common reference framework. \citet{snavely2008skeletal} partitioned extensive scenes by calculating the small bone skeleton set and reconstructing the skeleton set of the image. This approach reduces the number of parameters under consideration and enhances reconstruction efficiency. To sum up, global AAT offers high efficiency but suffers from poor robustness and scene integrity. On the other hand, incremental AAT exhibits high robustness and accuracy but tends to have relatively lower time efficiency. 

\subsection{Supervised Automated Aerial Triangulation}
Recognizing the limitations encountered by classical AAT algorithms, an increasing number of studies are exploring the application of deep learning methods to address these challenges. Many existing deep learning methods directly regress the depth map and pose of a monocular camera \citep{zhou2017unsupervised}, which usually highly rely on prior information for prediction. In addition, due to not considering the correlation between depth and pose, the generalization ability of these networks is limited, making it difficult to obtain ideal prediction results. BA-Net \citep{tang2018ba} attempts to use feature metric BA to solve the AAT problem. It makes end-to-end training possible by designing a differentiable LM (Levenberg–Marquardt) optimization algorithm, but the LM algorithm occupies a large amount of memory and has low computational efficiency. DeepSfM \citep{wei2020deepsfm} can simultaneously regress the pose and depth maps corresponding to the image; however, it requires coarse poses and depth maps for initialization and has high GPU requirements, making it difficult to scale up for high-resolution images and large-scale environments. DeepMLE \citep{xiao2022deepmle} does not require initial values as input, which expresses the two-view AAT problem as maximum likelihood estimation, learning the relative pose of the two views by maximizing the likelihood function of correlation. Similarly, for the problem of binocular vision estimation, \citet{wang2021deep} proposed a dense optical flow estimation network for predicting between two frames, which includes a scale-invariant depth estimation module that can simultaneously calculate the relative pose of the camera based on the corresponding relationship of 2D optical flow. DRO \citep{gu2021dro} is an optimization method based on recurrent neural networks that iteratively updates camera pose and image depth to minimize feature measurement errors. \citet{zhuang2021fusing} used a self-attention graph neural network to enhance strong interactions between different corresponding relationships and potentially model complex relationships between points to drive learning. MOAC \citep{wu2022moac} introduces a grouped dual cost enhancement module, which enhances the spatial semantic information and channel relationships of costs, making the optimization more robust to noise. \citet{moran2021deep} proposed a new approach to solve AAT problems using deep learning. They use matched feature points as input, and after permutation equivariant networks, predict the pose of each camera and 3D points in the scene. Compared to many existing deep learning AAT methods, it can be applied to large-scale reconstruction tasks in an unsupervised manner. However, there are mainly two drawbacks to it. The first one is that it cannot eliminate incorrectly matched point pairs, which means all the input pairs should be correct and is usually difficult to achieve. Another one is that its prediction results are still not satisfactory because of the limited generalizability.

In summary, most of the existing supervised methods can only handle a small number of low-resolution images, and their regression performance is also poor, lacking usability and practicality. Hence, the proposed DeepAAT addresses the existing challenges encountered by both classic and learning-based AAT algorithms, and presents a meticulously designed deep network tailored for UAV imagery, emphasizing efficiency, scene completeness, and practical applicability. The main contributions of this study are threefold:

(1) DeepAAT incorporates a spatial-spectral feature aggregation module, specifically combining both the spatial layout and spectral characteristics of an image set. This module boosts the network's ability to perceive the spatial arrangement of cameras and enhances the global regression capability for poses. 

(2) DeepAAT introduces an outlier rejection module according to global consistency, which effectively generates a reliability evaluation score for each feature correspondence. This approach facilitates the efficient and precise elimination of erroneous matching pairs, thereby ensuring accuracy and reliability throughout the entire 3D reconstruction process.

(3) DeepAAT can efficiently process hundreds of UAV images simultaneously, marking a significant breakthrough in enhancing the applicability of deep learning-based AAT algorithms. Furthermore, through a block fusion strategy, DeepAAT can be effectively scaled up for large-scale scenarios.

The rest of this paper is structured as follows. The preliminaries for our system are provided in Section \ref{section:pre}. A brief system overview including the hardware and software structure is provided in Section \ref{section:system}. A detailed description of DeepAAT is presented in Section \ref{section:deepaat} and experiments are conducted on UAV image datasets in Section \ref{section:exp}. Conclusion and future work are drawn in Section \ref{section:conclusion}.

\section{Preliminary \label{section:pre}}

\subsection{Problem Definition of Automated Aerial Triangulation}

The task of AAT refers to estimating the camera poses and 3D scene points corresponding to the 2D observations on the images. In classic photogrammetry, it is well studied and understood that the relative camera poses and 3D scene points can be solved only with the 2D observations \citep{he2018three}. The absolute camera poses and 3D scene points respectively to the geodesic framework can be then obtained with Ground Control Points (GCPs) or the GPS mounted on the UAV \citep{li2019nrli}.

Assume that the stationary targeting survey area is viewed by $M$ images, which are captured by the camera with known pre-calibrated intrinsic parameter $\mathbf{K}$ at different places along the UAV survey mission. The $M$ unknown camera poses are represented by a set of projection matrices $\mathcal{P}=\{\mathbf{P}_m|m=1,..., M\}$. Each projection matrix $\mathbf{P}_m$ with the size of $3\times4$ is constructed by camera rotation $\mathbf{R}_m \in SO(3)$ (corresponding quaternion is $\mathbf{q}_m$) and position $\mathbf{t}_m \in \mathbb{R}^3$ according to $\mathbf{P}_m=[\mathbf{R}_m|\mathbf{t}_m]$. Given $N$ 3D scene points in the targeting survey area $\mathcal{F}=\{\mathbf{F}_n|n=1,...,N\}$, each 3D scene point is written by $\mathbf{F}_n=[\mathbf{F}^1_n,\mathbf{F}^2_n,\mathbf{F}^3_n,1]^\top$ in homogeneous coordinates. If $\mathbf{F}_n$ can be observed by the $m^{th}$ image, its projection on the $m$'s image is given by Eq.\eqref{eq:projection}. As the depth information $\lambda_{m,n}$ is lost during the projection, $\mathbf{f}_{m,n}$ is an up-to-scale bearing vector.

\begin{equation}\label{eq:projection}
\mathbf{f}_{m,n}=[\mathbf{f}^1_{m,n},\mathbf{f}^2_{m,n},1]^\top = \frac{1}{\lambda_{m,n}} \mathbf{K}\mathbf{P}_m\mathbf{F}_n.
\end{equation}

In a typical AAT procedure, the initial step involves 2D feature detection and matching between pairwise images, which is carried out using the widely used Scale-Invariant Feature Transform (SIFT) \citep{lowe2004distinctive} or other robust feature detectors and descriptors \citep{dusmanu2019d2}. This step is not the main focus of this work. Subsequently, a set of 2D feature tracks denoted as $\mathcal{T}=\{\mathbf{T}_n|n=1,..., N\}$, is used as the input for our algorithm. It should be noted that track $\mathbf{T}_n$ corresponds to the 3D feature $\mathbf{F}_n$ and is constructed by a set of 2D observations from different images using Eq.\eqref{eq:track}:

\begin{equation}\label{eq:track}
\mathbf{T}_n=\{\mathbf{f}_{m,n}|m\in \mathcal{J}_n\},
\end{equation}
where $\mathcal{J}_n$ denotes the set of images that can observe the 3D feature $\mathbf{F}_n$. Then the tracks could be used to recover the camera poses and 3D scene points like the existing incremental \citep{schonberger2016structure} or global \citep{moulon2013global} AAT strategies.

\subsection{Projective Factorization}

Despite the mainstream AAT methods listed above, projective factorization \citep{sturm1996factorization} is also a long-established method in AAT. We provide a brief introduction to projective factorization, as it forms the foundation for the operation of the proposed network. The complete image projections, namely the 2D feature tracks $\mathcal{T}$, can be gathered into a measurement matrix $\mathbf{W}_{\mathbf{mes}}$ in Eq.\eqref{eq:fraction_def}:

\begin{equation}\label{eq:fraction_def}
\begin{split}
\mathbf{W}_{\mathbf{mes}} &\equiv 
\left[ \begin{matrix}
	\lambda _{1,1}\mathbf{f}_{1,1}&		\lambda _{1,2}\mathbf{f}_{1,2}&		\cdots&		\lambda _{1,N}\mathbf{f}_{1,N}\\
	\lambda _{2,1}\mathbf{f}_{2,1}&		\lambda _{2,2}\mathbf{f}_{2,2}&		\cdots&		\lambda _{2,N}\mathbf{f}_{2,N}\\
	\vdots&		\vdots&		\ddots&		\vdots\\
	\lambda _{M,1}\mathbf{f}_{M,1}&		\lambda _{M,2}\mathbf{f}_{M,2}&		\cdots&		\lambda _{M,N}\mathbf{f}_{M,N}\\
\end{matrix} \right]
\\
& = \left[ \begin{array}{c}
	\mathbf{KP}_1\\
	\mathbf{KP}_2\\
	\vdots\\
	\mathbf{KP}_M\\
\end{array} \right]
\left[ \begin{array}{c}
	\mathbf{F}_1\\
	\mathbf{F}_2\\
	\vdots\\
	\mathbf{F}_N\\
\end{array} \right] ^{\top}.  
\end{split}
\end{equation}

If the 3D scene points in the targeting survey area $\mathcal{F}$ are observed by all the images, the camera poses and 3D scene points can be recovered using Singular Value Decomposition (SVD) of $\mathbf{W}_{\mathbf{mes}}$ \citep{sturm1996factorization}. For the common cases of missing observations, the SVD can be replaced with iterative methods \citep{magerand2017practical,dai2013projective}. However, these methods are usually too weak for AAT in the presence of outliers and noise \citep{iglesias2023expose}, and can not be directly applied to large-scale AAT. Nevertheless, the formulation of $\mathbf{W}_{\mathbf{mes}}$ provides an ideal way of keeping spatial correlation information for neural networks.

\subsection{Permutation Equivariant Layer}

Let $\mathbf{W}$ be a tensor with the shape of $M\times N \times D$, whose row represents the image index, the column represents the feature track index and the third dimension represents the feature index. Taking measurement matrix $\mathbf{W}_{\mathbf{mes}}$ in Eq.\eqref{eq:fraction_def} as an example, $\mathbf{W}_{\mathbf{mes}}$ can be rearranged with the shape of $M\times N \times2$ (the third dimension records the feature coordinates on the image plane) to serve as input for the neural network for the sake of convenience.  To recover the camera poses and 3D scene points using a deep neural network, we expect a particular layer can output the same results irrespective of the order of the camera poses or the feature tracks. This reordering problem could be solved using the Permutation Equivariant Layer (PEL)\citep{hartford2018deep}, which was first introduced by \citet{moran2021deep} to handle SfM problem exploring tensor's exchangeability. 

\begin{figure}[!t]
\centering
\includegraphics[width=0.96\linewidth]{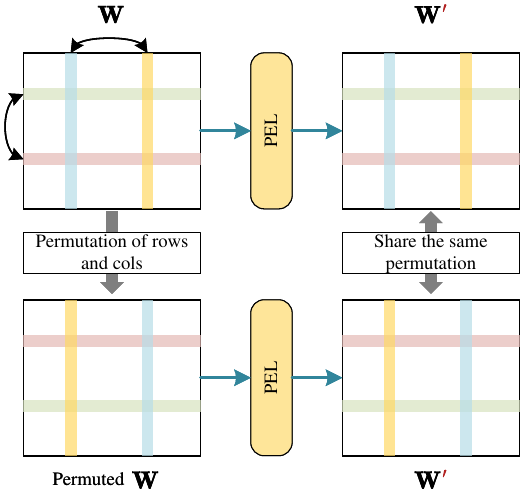}
\caption{ Exchangeability of tensor $\mathbf{W}$.}\label{fig:permutation}
\end{figure}

\textbf{Definition 1.} Exchangeability of a tensor $\mathbf{W}$ gives rise to the following property: If we permute arbitrary rows and columns of $\mathbf{W}$, then feed the permuted $\mathbf{W}$ into a PEL, the output tensor $\mathbf{W}^{\prime}$ should experience the same permutation of the rows and columns as illustrated in Fig.\ref{fig:permutation}.

\textbf{Theorem 1.} \citep{hartford2018deep} Take tensor $\mathbf{W}$ as input, the PEL with five unique parameters $h_1^{(d,o)},h_2^{(d,o)},...,h_4^{(d,o)}$ and $h_5^{(o)}$ could guarantee the output tensor  $\mathbf{W}^{\prime}$ with size of $M\times N \times O$ exchangeable following a specific fully connected layer calculation rule in Eq. \eqref{eq:permute_theorem}, where $d$ and $o$ are the indexes for the input and output feature channels, respectively.

\begin{figure*}[!t]
\begin{equation}\label{eq:permute_theorem}
\begin{split}
W_{m,n}^{\prime(o)}=\sigma \left( \sum_{d=1}^D{
\left( h_{1}^{\left( d,o \right)}W_{m,n}^{(d)} 
+\frac{h_{2}^{\left( d,o \right)}}{M}\left( \sum_{m^{\prime}}{W_{m^{\prime},n}^{(d)}} \right)
+\frac{h_{3}^{\left( d,o \right)}}{N}\left( \sum_{n^{\prime}}{W_{m,n^{\prime}}^{(d)}} \right) \right.} +\frac{h_{4}^{\left( d,o \right)}}{MN}\left( \sum_{m^{\prime},n^{\prime}}{W_{m^{\prime},n^{\prime}}^{(d)}} \right)
+h_{5}^{\left( o \right)} \right)  
\end{split}
\end{equation}
\end{figure*}
Inspired by the initial work proposed by \citet{moran2021deep}, we also utilize PEL to extract exchangeable high-level geometry correlations from the feature track matrix $\mathbf{W}$. But different from \citet{moran2021deep}, our proposed method takes into account not only the geometric features but also the spectral features of the feature tracks. Furthermore, the outliers in the feature track matrix are also automatically rejected to enhance the robustness of the results.

\section{System Overview}\label{section:system}

\begin{figure*}[!t]
\centering
\includegraphics[width=0.96\linewidth]{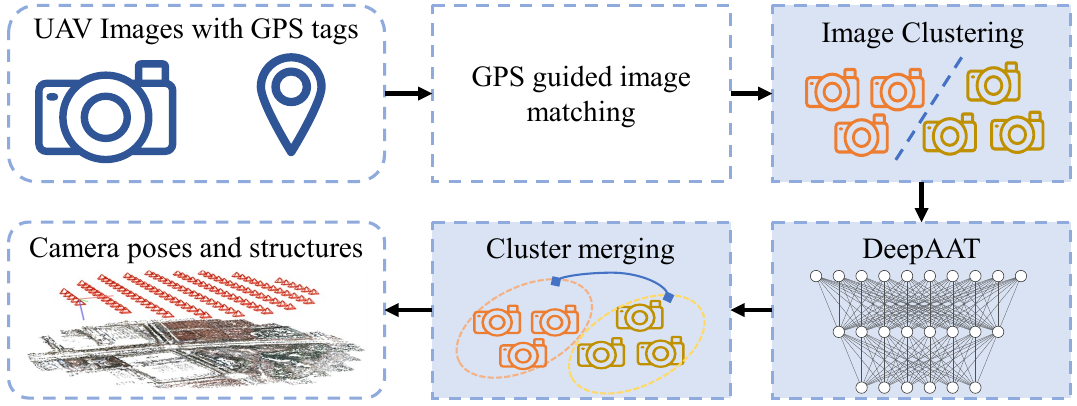}
\caption{ System overview of the efficient UAV-based mapping system.}\label{fig:system}
\end{figure*}

The proposed efficient UAV-based mapping system illustrated in Fig.\ref{fig:system} is briefly introduced in this section. As most UAV controllers, such as PixHawk \citep{meier2012pixhawk}, depend on GPS for trajectory planning and tracking in survey applications, it is assumed that each UAV image is geotagged with GPS information provided by the UAV controller. Despite the Single Point Positioning (SPP) error of GPS potentially reaching 10 meters on the UAV, it can still serve as a useful guide for the image matching process, focusing on matching nearby images only, as demonstrated by \cite{schonberger2016structure}. To be compatible with distributed parallel processing and limit the GPU memory usage on one computing unit for large-scale UAV-based mapping, the proposed system exploits the hierarchical SfM scheme \citep{chen2020graph,xu2021robust}, contains three components, namely, (1) \textbf{image clustering}, (2) \textbf{DeepAAT}, and (3) \textbf{cluster merging}.

(1) \textbf{Image clustering} divides the entire image set into multiple subsets considering the 2D feature correspondences between images. By treating the complete image set as a scene graph $\mathbf{G(V, E)}$ \citep{zhu2018very}, each image represents a vertex in $\mathbf{V}$, and an edge between two image vertices exists in $\mathbf{E}$ if the two images share feature correspondences. Setting the number of correspondences between images as the edge weight, $\mathbf{G(V, E)}$ is segmented using normalized cut \citep{shi2000normalized} iteratively until the number of images in each subset is within a desired number $N_{subset}$. $N_{subset}$ can be set according to the GPU memory on each computing unit. 

\textbf{Remark 1.} The number of 2D feature correspondences between pairs of images typically serves as a crucial metric for evaluating the reliability of relative matching. In essence, the greater the number of 2D feature correspondences between image pairs, the higher their matching reliability, and conversely, the fewer the matches, the lower the reliability. Our goal is to achieve a strong level of mutual matching within each subset. Consequently, the objective of the normalized cut operation on $\mathbf{G(V, E)}$ is to minimize the sum of edge weights within the cut, while also ensuring a balanced distribution of elements in each subset to enhance the computational efficiency for the following DeepAAT.

(2) \textbf{DeepAAT} efficiently and robustly recovers camera poses and structural information within each cluster. The network structure and implementation details of DeepAAT will be described in Section \ref{section:deepaat}.

(3) \textbf{Cluster merging} conducts a global bundle adjustment of the entire images taking the camera poses from each subset as initial values, hence recovering the whole camera poses and structures in the targeting survey area. More specifically, after rejecting the outlier tracks identified by the DeepAAT, the remaining feature tracks are then re-triangulated using the initial camera poses resulting from DeepAAT. Then the global bundle adjustment is performed once to get the final result. Readers can refer to \cite{triggs2000bundle} for additional details on re-triangulation and global bundle adjustment.

\section{Network Architecture of DeepAAT}\label{section:deepaat}

\begin{figure*}[!t]
\centering
\includegraphics[width=0.96\linewidth]{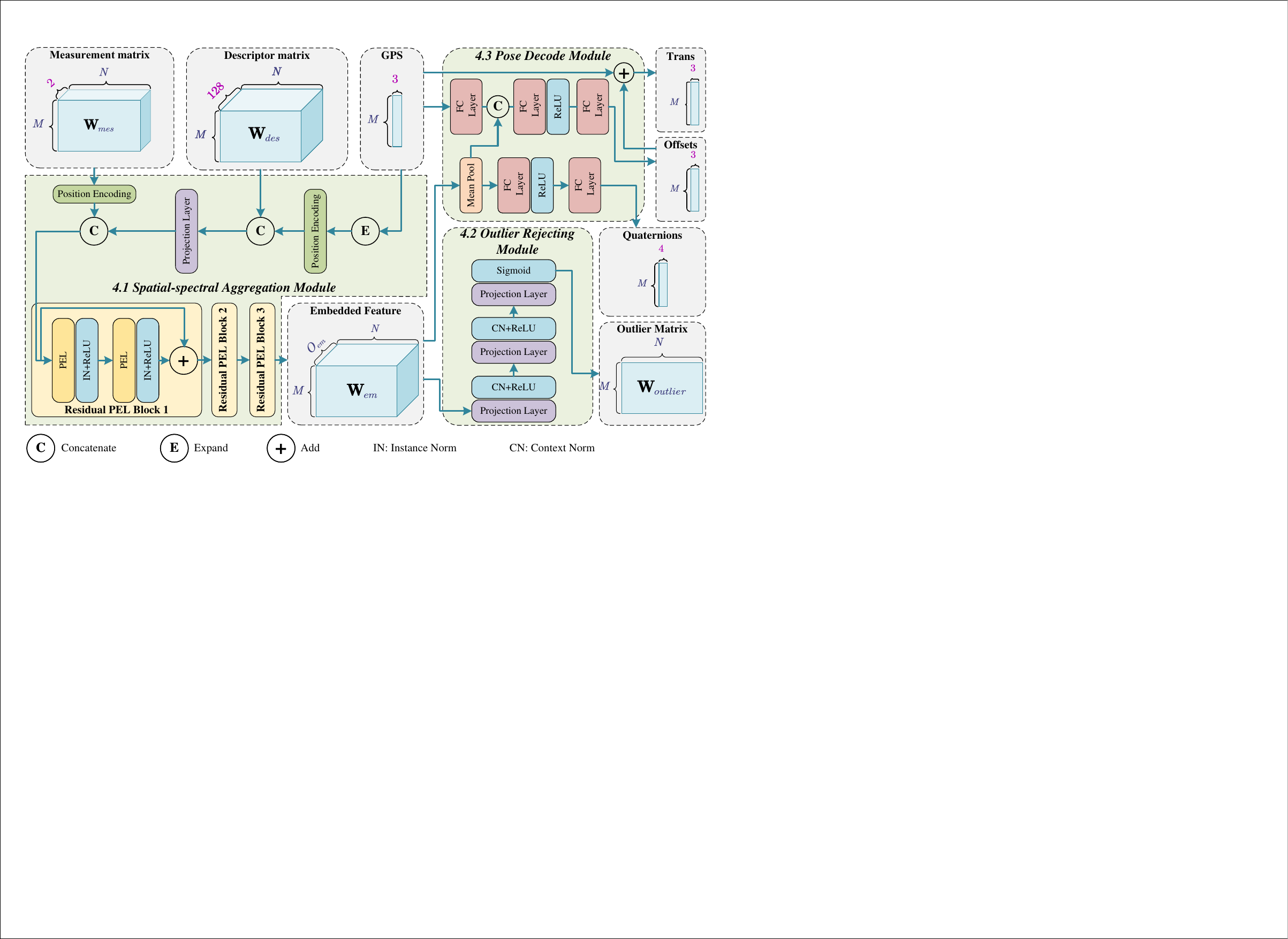}
\caption{ Network architecture of DeepAAT.}\label{fig:deepAAT}
\end{figure*}

The network architecture of DeepAAT mainly consists of three parts, spatial-spectral feature aggregation module (Section \ref{section:ssfa}), global consistency-based outlier rejecting module (Section \ref{section:gcor}), and pose decode module (Section \ref{section:pd}), which are illustrated in Fig.\ref{fig:deepAAT}. The input of DeepAAT includes feature measurement matrix $\mathbf{W}_{mes}$ constructed by reordering Eq.\eqref{eq:fraction_def} with the shape of $M\times N \times2$, SIFT feature descriptor matrix $\mathbf{W}_{des}$ with the shape of $M\times N \times 128$, and the geotag matrix $\mathbf{W}_{gps}$ with the shape of $M\times 3$. It's important to note that the GPS-derived latitude, longitude, and altitude values are initially transformed into the East-North-Up (ENU) local coordinate system  \citep{shin2002accuracy}. Following this transformation, they are further normalized to enhance network generativity. By feeding the input into the spatial-spectral feature aggregation module, a high-level embedded feature $\mathbf{W}_{em}$ is extracted for the downstream tasks. Then, the global consistency-based outlier rejecting module detects the outliers in the measurement matrix $\mathbf{W}_{mes}$ and gives the confidence for each 2D observation using $\mathbf{W}_{outlier}$. Meanwhile, the pose decode module recovers the camera poses. The loss functions will be detailed in Section \ref{section:loss}.

\subsection{Spatial-Spectral Feature Aggregation Module} \label{section:ssfa}

\textbf{Position Encoding:} The coordinates within the feature measurement matrix $\mathbf{W}_{mes}$ and the GPS matrix $\mathbf{W}_{gps}$ are both essential for the network to comprehend the spatial distribution. Position encoding \citep{mildenhall2021nerf}  has been employed for both of these location-related pieces of information to enhance the network's ability to distinguish relative positional relationships among input data, which is written as follows:
\begin{equation}\label{eq:position_encoding}
\footnotesize
\begin{split}
\varepsilon(x)=[sin(2^{0}\pi x),cos(2^{0}\pi x),...,sin(2^{L-1}\pi x),cos(2^{L-1}\pi x)]^\top,
\end{split}
\end{equation}
where $x$ is the coordinate value in each dimension, $L$ is the coding level. Position encoding solely influences the last dimension of $\mathbf{W}_{mes}$ and $\mathbf{W}_{gps}$. As for the GPS matrix $\mathbf{W}_{gps}$, to ensure its dimensional consistency with $\mathbf{W}_{mes}$ and $\mathbf{W}_{des}$, the proposed method conducts a dimension expansion operation in the first dimension (image indexes), transforming it from a two-dimensional matrix of \emph{M} × \emph{3} into a three-dimensional one of \emph{M} × \emph{N} × \emph{3}. Position encoding does not have learnable parameters, but it enhances the network's ability to distinguish location information.

\textbf{Residual Permutation Equivariant Layer:} The residual PEL employed in this paper consists of a consecutive pair of PEL, Instance Normalization (IN) \citep{ulyanov2016instance}, and Rectified Linear Unit (ReLU) combinations. Within the residual PEL, input and output data are summated through skip connections \citep{he2016deep}, serving two purposes: (1) ensuring stable gradient propagation within the network; and (2) facilitating the fusion of shallow layers, which contain more actual positional information, with deeper layers that offer more distinctive and discriminative feature information. It's worth noting that IN and ReLU do not alter the permutation equivariant properties of PEL.

\subsection{Global Consistency-based Outlier Rejecting Module} \label{section:gcor}

Even with pair-wise Epipolar geometry verification, $\mathbf{W}_{mes}$ still contain a substantial number of outlier matches, which can significantly impact obtaining correct global triangulation results and ensuring the proper convergence of the BA. Therefore, the proposed method utilizes a global consistency outlier rejecting module, which, by leveraging global information from the embedded feature $\mathbf{W}_{em}$, ensures that the subsequent BA operates successfully.

The global consistency-based outlier rejecting module primarily comprises three consecutive projection layers, which are fully connected (FC) layers that change the number of channels for non-empty data in sparse matrices, i.e. $Proj:\mathbb{R}^d\rightarrow\mathbb{R}^{d'}$. Following the first two projection layers are the Context Normalization (CN) and ReLU activation functions. The CN primarily serves to integrate data, allowing the previous layer's output to acquire corresponding context information in both camera poses and feature tracks. This aids the network in identifying outliers. Following the last projection layer is the Sigmoid. After passing through the Sigmoid function, the network's output is a score matrix with values ranging from 0 to 1, with dimensions $M \times N$. For each non-empty point, the score represents the probability that each 2D feature is an inlier. The closer the score is to 1, the higher the confidence that it is an inlier. In the outlier detection process, with a given threshold $\tau_{outlier}$, the predicted scores greater than $\tau_{outlier}$ are considered as inliers, while scores less than $\tau_{outlier}$ are considered outliers.

\subsection{Pose Decode Module} \label{section:pd}

The pose decode module utilizes the global information encoded by the spatial-spectral feature aggregation module to decode the camera's position and orientation. The decoder first performs mean pooling on the input feature along the column dimension (feature tracks), which maps the embedded feature $\mathbf{W}_{em}$ with the shape of $M \times N \times O_{em}$ to $M \times O_{em}$. The reason for choosing mean pooling is that each camera observes a different number of 3D points in the input scene. By averaging the features observed by each camera, the network can fairly represent the general characteristics of the scene observed by the $M$ cameras, regardless of the number of 3D points observed. This enables the decoder to obtain the global context information for each camera within the scene.

\textbf{Camera position decoder:} In the camera position decoding branch, GPS location information is reintroduced to improve the decoder's spatial positioning awareness. Additionally, the network performs regression on the camera's position offsets, which reflect the errors in the GPS tags, other than the camera's position directly. This is done because the magnitudes of GPS errors in each direction consistently fall within a specific range.

\textbf{Camera rotation decoder:} In the camera rotation decoding branch, the quaternion of each camera is regressed with two perception layers.

\subsection{Loss Function} \label{section:loss}

The loss function for DeepAAT comprises three components: $\mathcal{L}_{outlier}$, $\mathcal{L}_{position}$, and $\mathcal{L}_{rotation}$, which are written by:

\begin{equation}\label{eq:loss}
\begin{split}
\mathcal{L}=\mathcal{L}_{outlier}+\alpha \mathcal{L}_{position}+\beta \mathcal{L}_{rotation},
\end{split}
\end{equation}
where $\alpha$ and $\beta$ are balance factors. $\mathcal{L}_{outlier}$ is a Binary Cross-Entropy (BCE) like loss used to supervise the global consistency-based outlier rejecting module:

\begin{figure*}[!t]
\begin{equation}\label{eq:bce}
\begin{split}
\mathcal{L}_{outlier} = -\frac{1}{MN}\sum_{m=0}^{M-1}{\sum_{n=0}^{N-1}{\big(\mathbf{W}^{m,n}_{outlier}\mathbf{log}(\hat{\mathbf{W}}^{m,n}_{outlier}) + (1-\mathbf{W}^{m,n}_{outlier})\mathbf{log}(1-\hat{\mathbf{W}}^{m,n}_{outlier})}\big)},
\end{split}
\end{equation}
\end{figure*}

where $\hat{\mathbf{W}}^{m,n}_{outlier}$ is the predicted outlier score range from $[0,1]$. Both the rotation loss $\mathcal{L}_{rotation}$ and translation loss $\mathcal{L}_{position}$ are implemented using the mean square loss function:
\begin{equation}\label{eq:bce}
\begin{split}
\mathcal{L}_{rotation} = \frac{1}{M}\sum_{m=0}^{M-1}{||\mathbf{q}_m-\hat{\mathbf{q}}_m||_{2}},\\
\mathcal{L}_{position} = \frac{1}{M}\sum_{m=0}^{M-1}{||\mathbf{t}_m-\hat{\mathbf{t}}_m||_{2}},
\end{split}
\end{equation}
where $\hat{\mathbf{q}}_m$ and $\hat{\mathbf{t}}_m$ are the predicted rotation and translation for the $m^{th}$ camera.
\section{Experiments}\label{section:exp}

\subsection{Dataset}

\begin{figure*}[!t]
\centering
\includegraphics[width=1.0\linewidth]{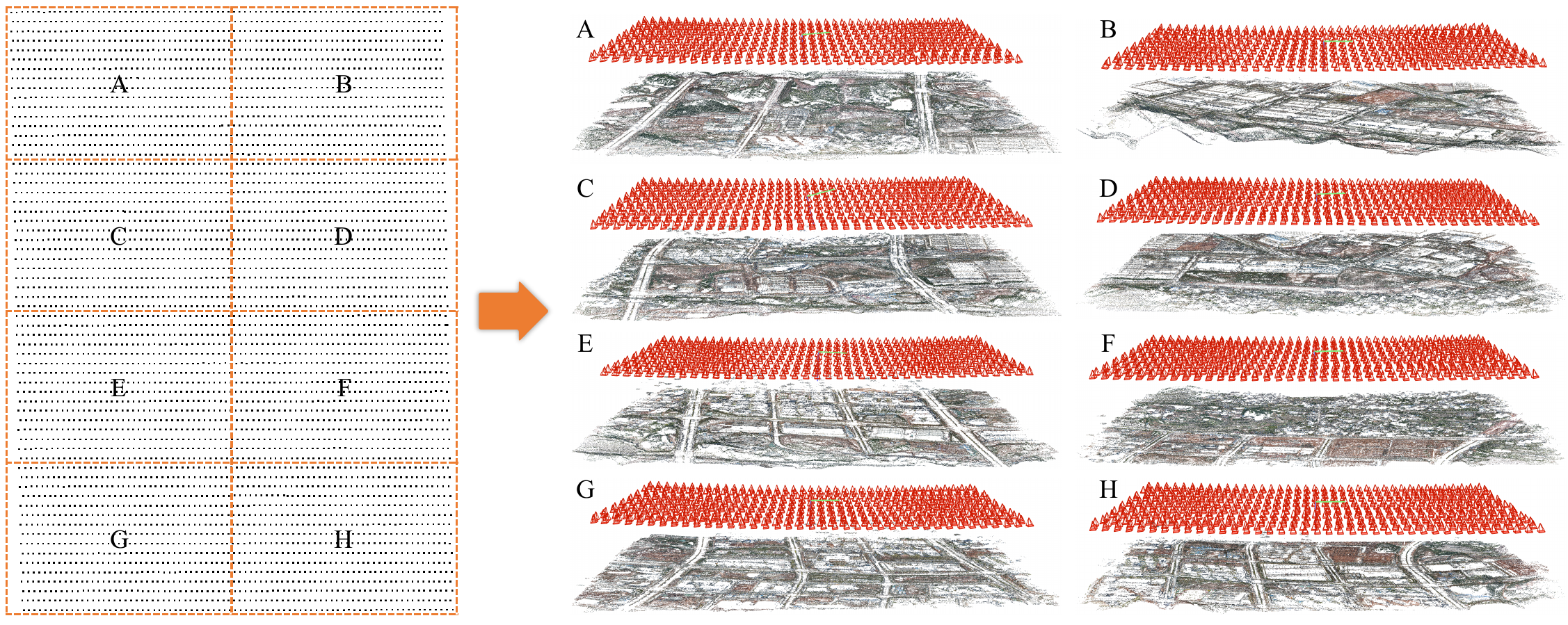}
\caption{ UAV-based image dataset used for the experiments. The dataset is divided into eight blocks.}\label{fig:dataset}
\end{figure*}

The experimental data, as depicted in Fig.\ref{fig:dataset}, was collected from an urban area including complex road networks, hills, construction sites, and buildings. A total of 4,992 UAV images were employed, subdivided into eight uniformed blocks labeled A to H. These images underwent preprocessing through SfM with high-precision Ground Control Points (GCPs) to establish reference data. Throughout the experiments, data from blocks A to G were used for training, while data from block H was employed as testing data. During dataset preparation, feature points that can be successfully matched but do not appear in the final AAT result are labeled as outliers.

\subsection{Implementation Details}
\subsubsection{Training sample generation}
The used training samples were generated through random sampling of images within each block. In the context of UAV AAT, typically, images that are closer to each other tend to have a greater number of feature correspondences and exhibit more stable connectivity. Specifically, for a given image set, one image is randomly selected to serve as the central image, denoted as $\mathbf{I}_c$. Then, according to the GPS position, Euclidean distances from all the other images to $\mathbf{I}_c$ are calculated and arranged in ascending order. Finally, based on the provided minimum and maximum sampling image limits, $N_{min}$ and $N_{max}$, a random number of sampled images, $N_{c}$, is determined. The $N_{c}$ closest images to $\mathbf{I}_c$ are selected to generate the sample data. Using this sampling strategy, a vast number of distinct samples can be generated. For instance, $N_{min}$ and $N_{max}$ are set as 100 and 130. A single image set can generate a total of $624 \times (N_{max}-N_{min}) = 18,720$ samples. 

\subsubsection{Data augmentation}
To enhance the network's generalizability, data augmentation is applied to the training data, focusing on two main aspects. Firstly, Gaussian noise with a mean of zero and a standard deviation of 0.01 was added to the input $\mathbf{W}_{mes}$. Secondly, given the limited amount of training data, and to augment and leverage the network's permutation equivariance capability, random row and column permutations were applied to the training samples in advance.

\subsubsection{Sparse matrix}
Because the number of scene points that can be observed in each image is limited, there will be a large number of zero elements in the measurement matrix $\mathbf{W}_{mes}$ and the descriptor matrix $\mathbf{W}_{des}$. Therefore, these matrices are implemented in the form of sparse matrices in the code to improve the processing efficiency of the network.

\subsubsection{Parameter settings}
The experiments involve the configuration of certain hyperparameters. Specifically, regarding position encoding, the encoding order is set at $L=4$. In the spatial-spectral aggregation module, the embedded feature dimension is set at $O_{em}=256$. In the outlier rejecting module, the outlier detection threshold is set at $\tau_{outlier}=0.5$. In the configuration of weights within the loss function, the parameters $\alpha$ and $\beta$, which govern the weights for rotation and translation, are assigned values of 0.9 and 0.1, respectively. This allocation stems from the observation that, in contrast to directly predicted rotation, the prediction of translation is effectively an adjustment relative to the initial estimate. Consequently, it is appropriate to assign a lesser weight to translation compared to rotation.

\subsubsection{Evaluation criteria}
This paper evaluates the experimental results from various perspectives. For the outlier removal results, evaluation metrics commonly used in binary classification tasks are employed, including $Accuracy$, $Precision$, $Recall$, and $F_1$ score, which are calculated as follows:

\begin{equation}\label{eq:prf}
\begin{split}
Accuracy &= \frac{TP+TN}{TP+FP+TN+FN},\\
Precision &= \frac{TP}{TP+FP},\\
Recall &= \frac{TP}{TP+FN},\\
F_1 &= \frac{2\times Precision \times Recall}{Precision + Recall},
\end{split}
\end{equation}
where $TP$ is True Positive, $FP$ is False Positive, $TN$ is True Negative, $FN$ is False Negative. 

For the reconstruction results, we use the reprojection error $E_{repro}$, position error $E_t$ and angle error $E_R$ to evaluate from three aspects.

\begin{equation}\label{eq:prf}
\begin{split}
\left\{ \begin{matrix}
{E_{repro} = \frac{1}{n_{2d}}{\sum\limits_{i = 1}^{m}{\sum\limits_{j = 1}^{n}\left\| \left( {x_{ij}^{1} - \frac{P_{i}^{1}X_{j}}{P_{i}^{3}X_{j}},x_{ij}^{2} - \frac{P_{i}^{2}X_{j}}{P_{i}^{3}X_{j}}} \right) \right\|_{2}}}}, \\
{E_{t} = \frac{1}{m}{\sum\limits_{i = 1}^{m}\left\| {{\hat{t}}_{i} - {\overset{\sim}{t}}_{i}} \right\|_{2}}}, \\
{E_{R} = \frac{1}{m}{\sum\limits_{i = 1}^{m}{\cos^{- 1}\left( {\frac{1}{2}\left( {tr\left( {{\hat{R}}_{i}^{T}{\overset{\sim}{R}}_{i}} \right) - 1} \right)} \right)}}},
\end{matrix} \right.
\end{split}
\end{equation}
where $n_{2d}$ represents the number of 2D pixels in the scene, $m$ is the number of cameras, $n$ is the number of 3D points, and $x^k_{ij}$ represents the $k$th dimension of the coordinate of the $j$th 3D point observed by the $i$th image, $P_i^k$ denotes the $k$th row of the $i$th camera matrix, $X_j$ denotes the coordinate of the $j$th 3D point, $\hat{t}_i$ is the reference value of camera position, $\widetilde{t}_i$ is the predicted camera position, $\hat{R}_i$ is the reference value of camera rotation, $\widetilde{R}_i$ is the predicted camera rotation value, and $tr()$ denotes the trace of the matrix (i.e. the sum of the main diagonal elements of the matrix).

In addition, we also record the time used in network prediction and compare it with baseline methods as an important evaluation indicator.

\subsubsection{Computation resources}
The configuration of the machine used in our experiments is as follows. CPU: Intel (R) Xeon (R) Silver 4210R CPU @ 2.40GHz, GPU: NVIDIA A100 SXM4 80GB. To control the memory size and computational complexity, all network training and prediction tasks involved in this article can be run on a single Tesla V100 with 32GB memory.

\subsection{Results of scene segmenting and merging }
\begin{figure}[!t]
\centering
\includegraphics[width=0.96\linewidth]{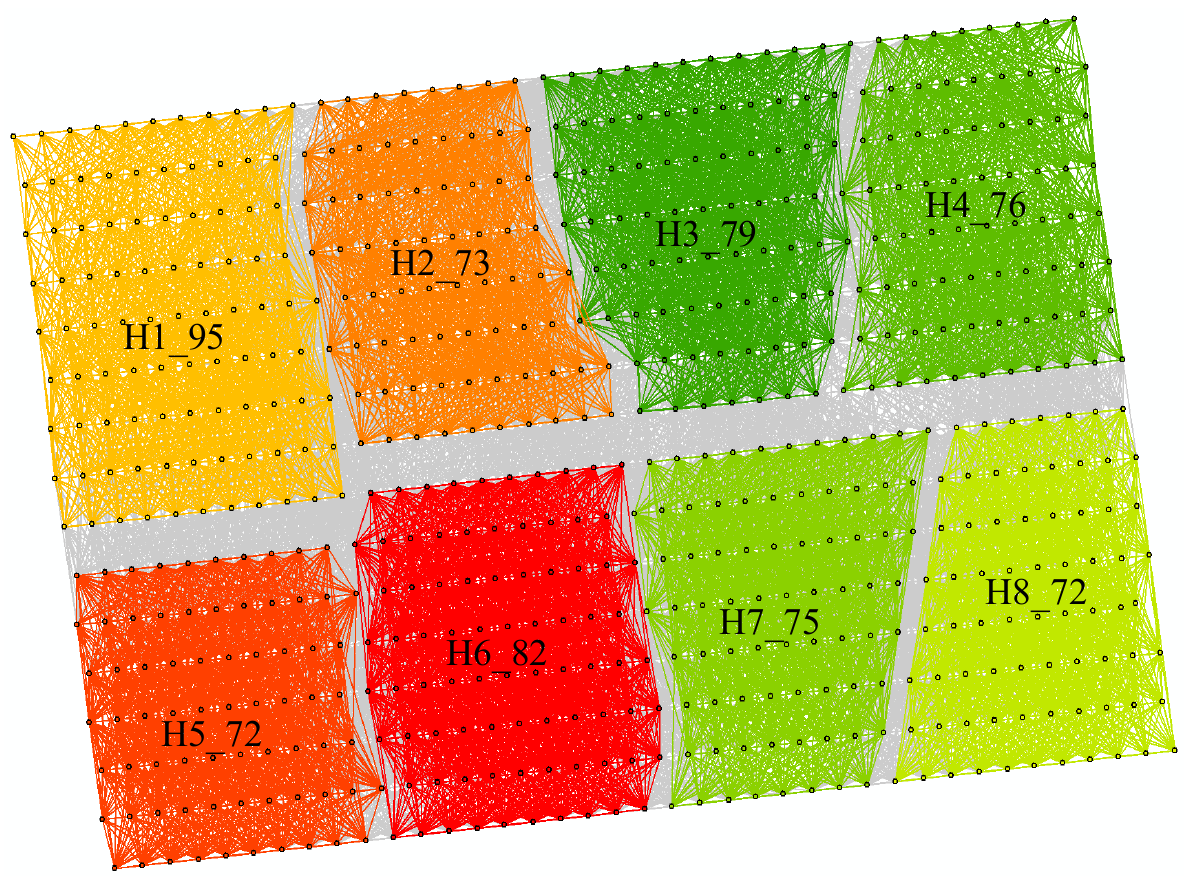}
\caption{ Image clustering result, where the number after "\_" represents the number of cameras included in the subset.}\label{fig:clustering}
\end{figure}

To facilitate large-scale reconstruction tasks, we employed a strategy of image clustering and merging. During the image clustering phase, we set a maximum limit of 100 cameras for each subset. Consequently, block H was ultimately divided into 8 distinct blocks, with each subset containing between 72 to 95 cameras. The clustering outcome is illustrated in Fig.\ref{fig:clustering}, where each circle denotes a camera, and different colors represent individual subsets. Notably, the gray lines in the figure are the severed edges that connect cameras across different subsets.

\begin{table}[]
\caption{Outlier rejection result}\label{table:orr}
\centering
\normalsize
\resizebox{0.5\textwidth}{!}{
\begin{tabular}{c c c c c c}
\hline
Scene & ↑Acc & ↑Pre & ↑Rec & ↑F1 & Time/s\\
\hline
H1\_95&0.959&0.966&0.980&0.973&0.820\\
H2\_73&0.933&0.960&0.952&0.956&0.087\\
H3\_79&0.947&0.966&0.968&0.967&0.094\\
H4\_76&0.951&0.960&0.974&0.967&0.085\\
H5\_72&0.967&0.978&0.982&0.980&0.117\\
H6\_82&0.965&0.980&0.979&0.980&0.129\\
H7\_75&0.952&0.975&0.966&0.970&0.092\\
H8\_72&0.967&0.981&0.980&0.980&0.080\\
Mean&0.955&0.971&0.973&0.972&0.188\\
\hline
\end{tabular}
}
\begin{tablenotes}
\footnotesize
\item *Acc(Accuracy), Pre(Precision), Rec(Recall), F1(F1 Score)
\end{tablenotes}
\end{table}

Table \ref{table:orr} shows that the average performance of the outlier rejection in the scene surpasses 0.95 across all four metrics, with a notable recall rate of 0.973. This indicates that about 97.3\% of the pixels identified by the network as positive are indeed true positives. Such high accuracy is advantageous for the subsequent steps of global triangulation and BA. This also demonstrates that the global consistency-based outlier rejection module, as designed in this paper, is highly effective and applicable throughout the entire algorithmic process. Moreover, for clustered scenes, the network's prediction time is under one second, highlighting the proposed network's high operational efficiency in AAT tasks. Here, we predict all 8 scenarios through a single model loading, and except for the first scenario, the remaining 7 scenarios do not require reloading the model, resulting in a nearly tenfold reduction in time consumption.

\begin{table}[]
\caption{Pose prediction result}\label{table:ppr}
\centering
\resizebox{0.5\textwidth}{!}{
\begin{tabular}{cccccccc}
\hline
\multirow{2}{*}{Scene} & \multirow{2}{*}{IPE/m} & \multicolumn{3}{c}{DeepAAT}                                    & \multicolumn{3}{c}{Results after BA}                           \\ \cline{3-8} 
                       &                                 & RPE/pix & PE/m & RE/° & RPE/pix & PE/m & RE/° \\ \hline
H1\_95                 & 5.153                           & 64.495                   & 4.170            & 1.961            & 0.490                    & 2.732            & 0.032            \\
H2\_73                 & 5.159                           & 52.222                   & 4.495            & 1.322            & 0.459                    & 1.865            & 0.048            \\
H3\_79                 & 5.192                           & 60.994                   & 4.941            & 1.845            & 0.475                    & 1.994            & 0.032            \\
H4\_76                 & 4.955                           & 52.681                   & 3.868            & 1.903            & 0.467                    & 2.570            & 0.023            \\
H5\_72                 & 5.259                           & 45.699                   & 4.184            & 1.155            & 0.487                    & 2.583            & 0.027            \\
H6\_82                 & 5.417                           & 59.702                   & 4.376            & 2.042            & 0.464                    & 1.938            & 0.027            \\
H7\_75                 & 5.262                           & 53.557                   & 4.530            & 1.897            & 0.477                    & 1.801            & 0.034            \\
H8\_72                 & 5.091                           & 45.062                   & 4.200            & 1.813            & 0.486                    & 2.375            & 0.028            \\
Mean                   & 5.186                           & 54.302                   & 4.346            & 1.742            & 0.476                    & 2.232            & 0.031            \\ \hline
\end{tabular}
}
\begin{tablenotes}
\footnotesize
\item *IPE(Initial Position Error), RPE(Reprojection Error), PE(Position Error), RE(Rotation Error)
\end{tablenotes}
\end{table}

\begin{figure*}[!t]
\centering
\includegraphics[width=1.0\linewidth]{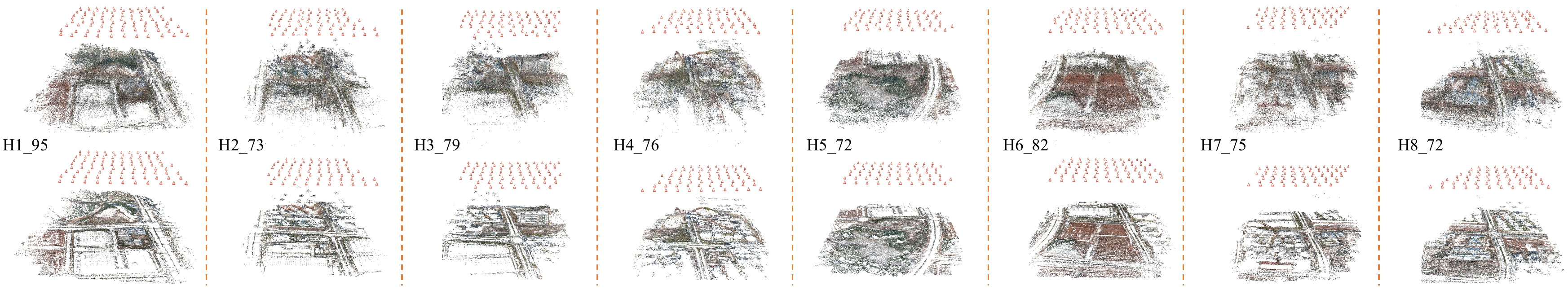}
\caption{ Network prediction results (upper) and results after BA (lower).}\label{fig:prediction}
\end{figure*}

Table \ref{table:ppr} demonstrates that across all eight clustered scenes, the predicted camera position error is consistently lower than the initial position error. The results show greater precision in predicting rotation, with an average error of less than 2°, which is highly beneficial for accurate subsequent adjustments. After BA, the average reprojection error across all scenes is less than 0.5 pixels, and the rotation error is under 0.1°. The visualized results, as depicted in Fig.\ref{fig:prediction}, further affirm the effectiveness of the algorithm.

\begin{figure}[!t]
\centering
\includegraphics[width=0.96\linewidth]{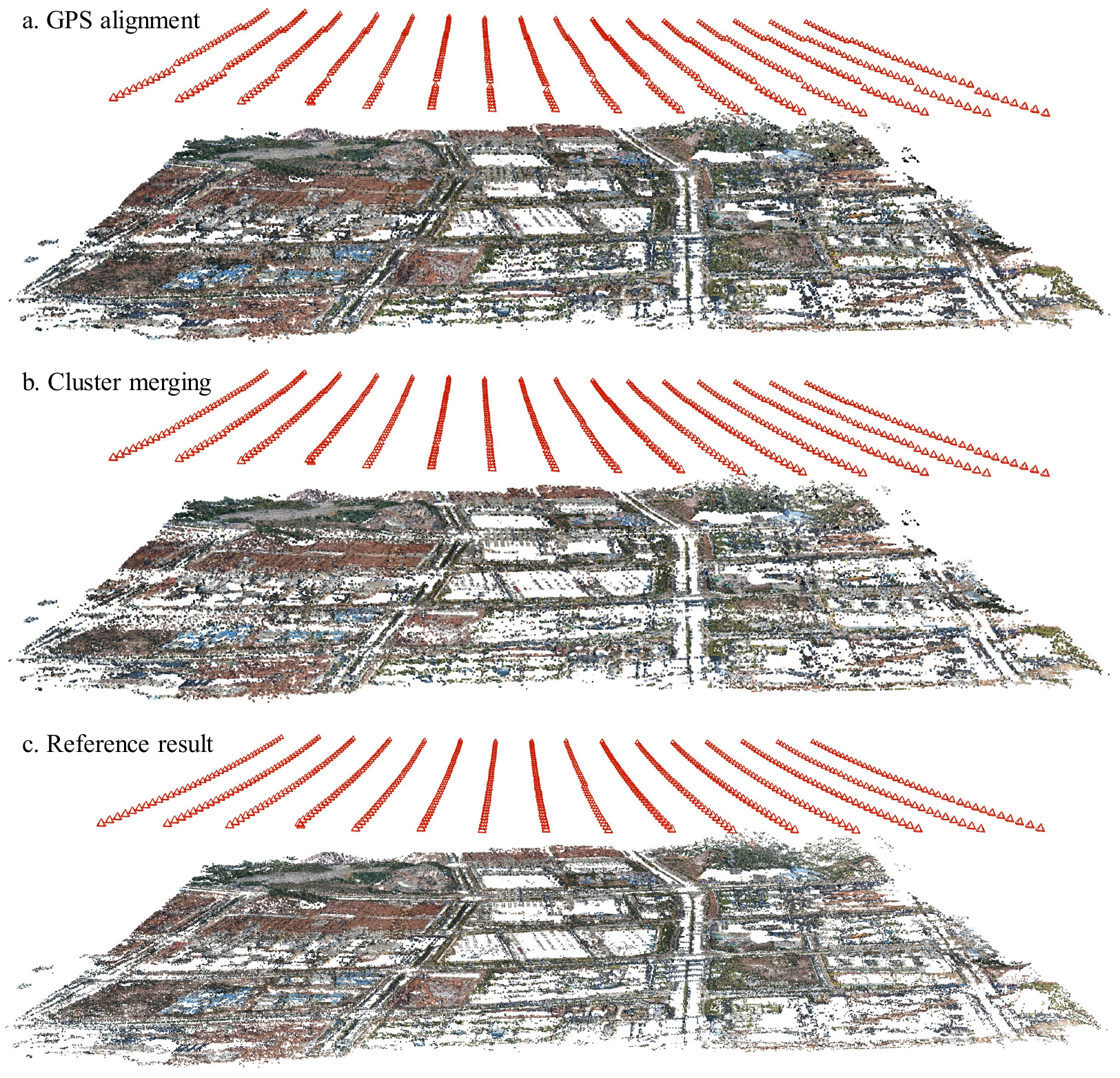}
\caption{ (a) GPS alignment result, (b) cluster merging result, and (c) reference result.}\label{fig:merging}
\end{figure}

In the following, the Cluster Merging algorithm described in Section \ref{section:system} is used to fuse the above 8 segmented scenes, and the results are shown in Fig.\ref{fig:merging} and Tab.\ref{table:cmr}. From Fig.\ref{fig:merging}, it can be seen that after using GPS information to globally align all segmented scenes, there is a significant offset between different scenes. After Cluster Merging, the inconsistency between scenes was effectively eliminated, resulting in globally consistent fusion results that were similar in appearance to the reference one. From Tab.\ref{table:cmr}, it can be seen that the reprojection error of the scene has slightly decreased compared to the average reprojection error of the segmented scene, but in terms of position error and rotation error, it has slightly increased compared to the average result of the segmented scene. In addition, the reconstructed scene points have a slight increase compared to the 3D points of the reference scene. These results indicate that the proposed hierarchical AAT scheme can effectively complete large-scale AAT tasks.

\begin{table}[]
\caption{Cluster merging result}\label{table:cmr}
\centering
\normalsize
\noindent
\begin{tabular}{c c c c c}
\hline
RPE/pix & PE/m & RE/° & Points & Reference points\\
\hline
0.471 & 2.726 & 0.450 & 553164 & 549517\\
\hline
\end{tabular}
\end{table}

\subsection{Comparison}
We compare the proposed algorithm with the SOTA methods, including:

ESFM \citep{moran2021deep}: A neural network architecture is proposed, which takes track points as input in the form of a matrix. It can simultaneously predict camera pose and scene points, and use reprojection error as the loss function.

Colmap \citep{schonberger2016structure}: A state-of-the-art open-source incremental SfM pipeline library, widely used in pose estimation and scene reconstruction tasks. Colmap provides both UI interface and command line running mode, making it easy to operate and has good reconstruction results.

OpenMVG \citep{moulon2013adaptive}: Provides both incremental SfM and global SfM implementations, with global SfM being the current SOTA in the open-source library. By using the command line, step-by-step SfM can be easily implemented.

\begin{figure*}[!t]
\centering
\includegraphics[width=0.96\linewidth]{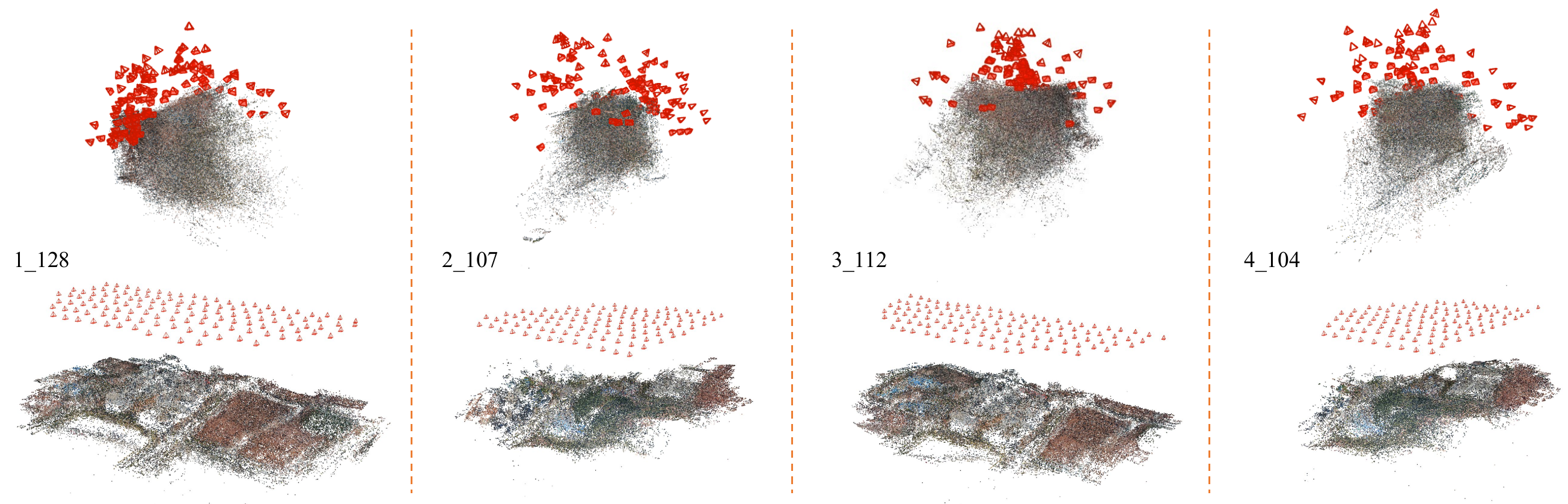}
\caption{ Comparison of partial scenarios predicted by ESFM (Upper) and DeepAAT (Lower).}\label{fig:comp_esfm}
\end{figure*}

Because the output of ESFM and the proposed method is the result before BA, we directly compared the predicted results of the two networks, including reprojection error and rotation error. The results are shown in Tab.\ref{table:comp_esfm}, and some prediction scenarios are shown in Fig.\ref{fig:comp_esfm}. To ensure consistency in the learning space of ESFM, we standardized all input scenes, allowing the network to learn the relative positions of all cameras with respect to the first camera in the scene. When comparing the proposed DeepAAT with Colmap and OpenMVG, we directly compared the final reconstruction results after BA, including reprojection error, final scene points, and time consumption. The results are shown in Tab.\ref{table:comp_trad}, and the predicted scenes are shown in Fig.\ref{fig:comp_woor}.

\begin{table}[]
\caption{Comparison of ESFM and the proposed DeepAAT}\label{table:comp_esfm}
\centering
\resizebox{0.5\textwidth}{!}{
\begin{tabular}{cccccccccccc}
\cline{1-7}
\multirow{2}{*}{Scene} & \multicolumn{3}{c}{ESFM}                                          & \multicolumn{3}{c}{DeepAAT (ours)}\\ \cline{2-7}
                       & ↓RPE/pixel & ↓PE/m & ↓RE/° & ↓RPE/pixel & ↓PE/m & ↓RE/° \\ \cline{1-7}
1\_128                 & 410.580                   & 229.447           & 62.265            & \textbf{46.891}                    & \textbf{4.340}             & \textbf{1.610} \\
2\_107                 & 460.225                   & 207.427           & 103.297           & \textbf{58.965}                    & \textbf{3.807}             & \textbf{2.409}  \\
3\_112                 & 356.813                   & 225.512           & 55.177            & \textbf{46.582}                    & \textbf{3.995}             & \textbf{1.422}  \\
4\_104                 & 442.035                   & 209.172           & 101.788           & \textbf{58.436}                    & \textbf{4.301}             & \textbf{2.363}  \\
5\_104                 & 440.410                   & 197.034           & 92.420            & \textbf{41.844}                    & \textbf{3.962}             & \textbf{2.091}  \\
6\_106                 & 457.345                   & 204.852           & 59.944            & \textbf{49.994}                    & \textbf{4.087}             & \textbf{1.713}  \\
7\_118                 & 397.622                   & 205.955           & 60.104            & \textbf{43.848}                    & \textbf{3.957}             & \textbf{1.631}  \\
8\_127                 & 433.450                   & 215.331           & 68.668            & \textbf{45.161}                    & \textbf{4.177}             & \textbf{1.417}  \\
9\_127                 & 362.216                   & 220.011           & 95.587            & \textbf{49.067}                    & \textbf{4.455}             & \textbf{1.676}  \\
10\_129                & 479.165                   & 223.690           & 66.945            & \textbf{44.770}                    & \textbf{4.144}             & \textbf{1.483}  \\
mean                   & 423.986                   & 213.843           & 76.619            & \textbf{48.556}                    & \textbf{4.123}             & \textbf{1.781}  \\ \cline{1-7}
\end{tabular}}
\end{table}

Tab.\ref{table:comp_esfm}, indicates that the prediction results of the proposed DeepAAT are much smaller than ESFM in terms of reprojection error, position error, and rotation error. From the four comparative scenarios in Fig.\ref{fig:comp_esfm}, it can be seen that the scenarios predicted by the proposed DeepAAT are all correct, but the prediction results of ESFM are relatively chaotic, resulting in the inability to obtain correct results through subsequent global BA. The main reasons are as follows:

(1) By integrating GPS information as prior knowledge, DeepAAT significantly enhances its spatial awareness and perception of location. Crucially, DeepAAT predicts the offset in camera position, which is a relative measure to GPS coordinates, rather than attempting to directly ascertain the precise location of each camera.

(2) In contrast to ESFM, DeepAAT incorporates a global consistency-based outlier rejection module, which effectively eliminates erroneous matching points that persist even after geometric verification. As a result, the prediction outcomes produced by DeepAAT are considerably more refined and cleaner. In contrast, ESFM lacks a denoising feature, and the presence of noise points in its framework can adversely affect the network's ability to accurately learn and represent the correct scene.

\begin{table*}[]
\caption{Comparison results between DeepAAT and traditional algorithms}\label{table:comp_trad}
\centering
\resizebox{0.96\textwidth}{!}{
\begin{tabular}{ccccccccc}
\hline
\multirow{2}{*}{Scene} & \multicolumn{2}{c}{Colmap}            & \multicolumn{2}{c}{OpenMVG Incremental}      & \multicolumn{2}{c}{OpenMVG Global}    & \multicolumn{2}{c}{DeepAAT (Ours)}              \\ \cline{2-9} 
                       & ↓RPE/pix & ↓Time /s & ↓RPE/pix & ↓Time/s    & ↓RPE/pixel & ↓Time/s & ↓RPE/pix & ↓Time/s \\ \hline
1\_128                 & 0.500                      & 465.966  & \textbf{0.478}                          & 548.536     & 0.548                      & 29.651   & 0.489                      & \textbf{0.845}    \\
2\_107                 & 0.566                      & 296.569  & 0.518                          & 390.071     & 0.611                      & 15.830   & \textbf{0.482}                      & \textbf{0.832}    \\
3\_112                 & 0.501                      & 377.211  & 0.479                          & 601.795     & 0.539                      & 28.098   & \textbf{0.472}                      & \textbf{0.861}    \\
4\_104                 & 0.554                      & 276.811  & 0.522                          & 418.626     & 0.611                      & 14.442   & \textbf{0.491}                      & \textbf{0.798}    \\
5\_104                 & 0.528                      & 285.427  & 0.473                          & 367.106     & 0.590                      & 13.879   & \textbf{0.447}                      & \textbf{0.780}    \\
6\_106                 & 0.506                      & 365.869  & \textbf{0.450}                          & 417.672     & 0.516                      & 19.331   & 0.484                      & \textbf{0.815}    \\
7\_118                 & 0.507                      & 375.640  & 0.478                          & 564.906     & 0.550                      & 25.276   & \textbf{0.467}                      & \textbf{0.831}    \\
8\_127                 & 0.521                      & 437.120  & \multicolumn{2}{c}{Reconstruction   failure} & 0.548                      & 34.464   & \textbf{0.473}                      & \textbf{0.862}    \\
9\_127                 & \textbf{0.478}                      & 439.329  & 0.481                          & 423.064     & 0.551                      & 22.412   & 0.483                      & \textbf{0.859}    \\
10\_129                & 0.503                      & 465.138  & \textbf{0.458}                          & 615.676     & 0.522                      & 28.488   & 0.487                      & \textbf{0.868}    \\
Mean                   & 0.516                      & 378.508  & 0.482                          & 484.050     & 0.559                      & 23.187   & \textbf{0.478}                      & \textbf{0.835}    \\ \hline
\end{tabular}}
\end{table*}


Table \ref{table:comp_trad} reveals that DeepAAT outperforms all other methods in terms of average reprojection error and average time consumption, with both indicators surpassing those of Colmap and OpenMVG. Its most striking advantage lies in time efficiency, as DeepAAT significantly outpaces the comparison methods across all test scenarios. Specifically, DeepAAT's average reconstruction efficiency is 453 times greater than Colmap, 580 times that of OpenMVG Incremental, and 28 times that of OpenMVG Global. This suggests that the proposed network substantially enhances the efficiency of AAT reconstruction while maintaining scene integrity. Concurrently, it also enhances reconstruction accuracy to a certain extent. As indicated in Table \ref{table:comp_trad}, OpenMVG Incremental failed to reconstruct scene 8\_127, likely due to the stringent requirements of the incremental SfM algorithm on initial image pair selection and the relative instability of the OpenMVG Incremental algorithm.

\subsection{Ablation Study}
To test the impact of the core modules proposed in DeepAAT, the following ablation experiments were conducted: \textcircled{1} The encoding order of GPS $L_G$ and the measurement matrix $L_{mes}$ is set to 2; \textcircled{2} The encoding order of GPS $L_G$ and the measurement matrix $L_{mes}$ is set to 4, which is the setting used in this article; \textcircled{3} The encoding order of GPS $L_G$ and the measurement matrix $L_{mes}$ is set to 6; \textcircled{4} Remove the spatial spectral feature aggregation module, similar to ESFM, using only the measurement matrix $\mathbf{W}_{mes}$ as the network input; \textcircled{5} Remove the global consistency-based outlier rejecting module. Here, the predicted pose of the network is directly used to triangulate all matching track points during global triangulation. These experimental data are the average of 10 test data results, which are shown in Tab.\ref{table:adstudy}. From Tab.\ref{table:adstudy}, it can be seen that:

\begin{table}[]
\caption{Results of ablation study}\label{table:adstudy}
\centering
\resizebox{0.5\textwidth}{!}{
\begin{tabular}{cccccccc}
\hline
\multirow{2}{*}{Setting} & \multicolumn{4}{c}{Outlier rejection}        & \multicolumn{3}{c}{Pose estimation}                               \\ \cline{2-8} 
                         & ↑Acc & ↑Pre & ↑Rec & ↑F1 & ↓RPE/pix & ↓PE/m & ↓RE/° \\ \hline
\textcircled{1}                      & \textbf{0.967}     & 0.973      & 0.987   & \textbf{0.980}     & 49.460                    & \textbf{3.974}             & 2.411             \\
\textcircled{2}                      & 0.966     & \textbf{0.974}      & 0.985   & 0.979     & \textbf{48.556}                    & 4.123             & \textbf{1.781}             \\
\textcircled{3}                      & \textbf{0.967}     & 0.971      & \textbf{0.989}   & \textbf{0.980}     & 57.947                    & 4.953             & 2.036             \\
\textcircled{4}                      & 0.964     & 0.970      & 0.986   & 0.978     & 98.414                    & 5.104             & 2.990             \\
\textcircled{5}                      & /         & /          & /       & /         & 48.599                    & 4.123             & 1.781             \\ \hline
\end{tabular}}
\end{table}

(1) Encoding order $L$ of GPS information and the measurement matrix in the spatial-spectral feature aggregation module has little impact on the experimental results, indicating that as long as the encoding order is set within a reasonable range, good experimental results can be achieved.


(2) Upon removal of the spatial-spectral feature aggregation module, there was a marked decline in the network's overall performance, particularly notable in pose prediction tasks. The reprojection error more than doubled compared to the optimal outcome, accompanied by substantial deviations in both position and rotation errors. These experimental findings highlight that the integration of GPS positioning data and feature point descriptor into the network input plays a critical role in significantly enhancing network performance.

\begin{figure*}[!t]
\centering
\includegraphics[width=1\linewidth]{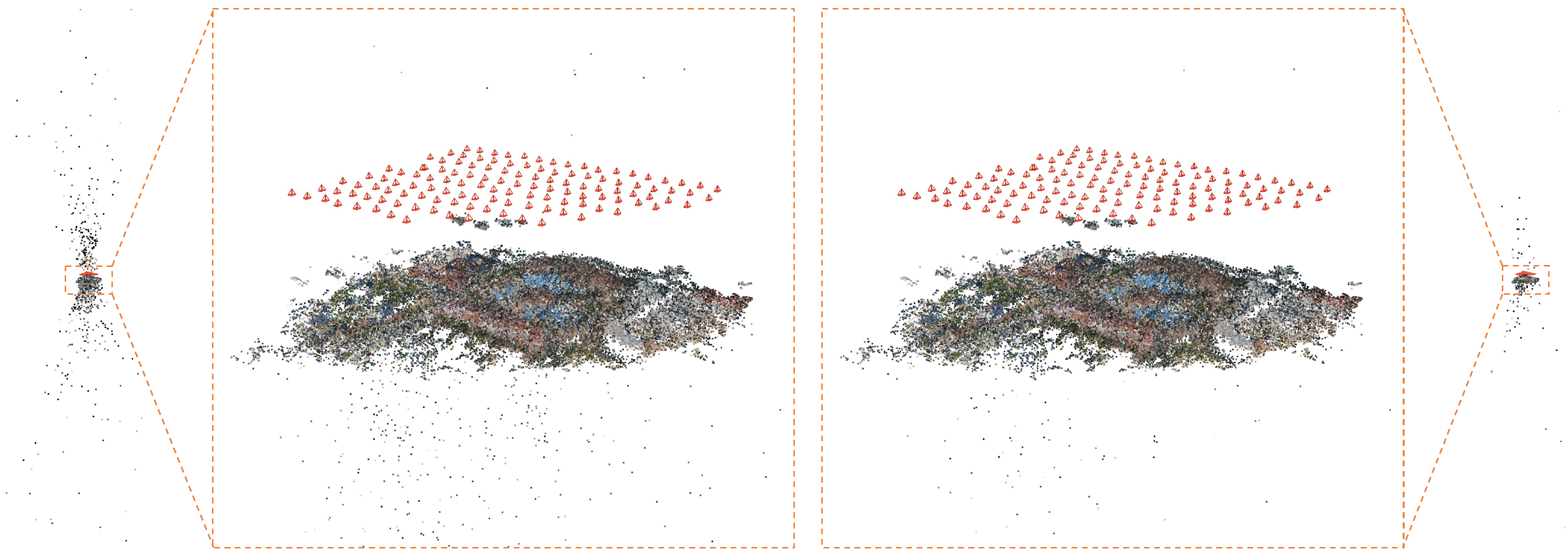}
\caption{ Prediction Scenarios without outlier rejecting (Left) and with outlier rejecting (Right).}\label{fig:comp_woor}
\end{figure*}
\begin{figure}[!t]
\centering
\includegraphics[width=0.96\linewidth]{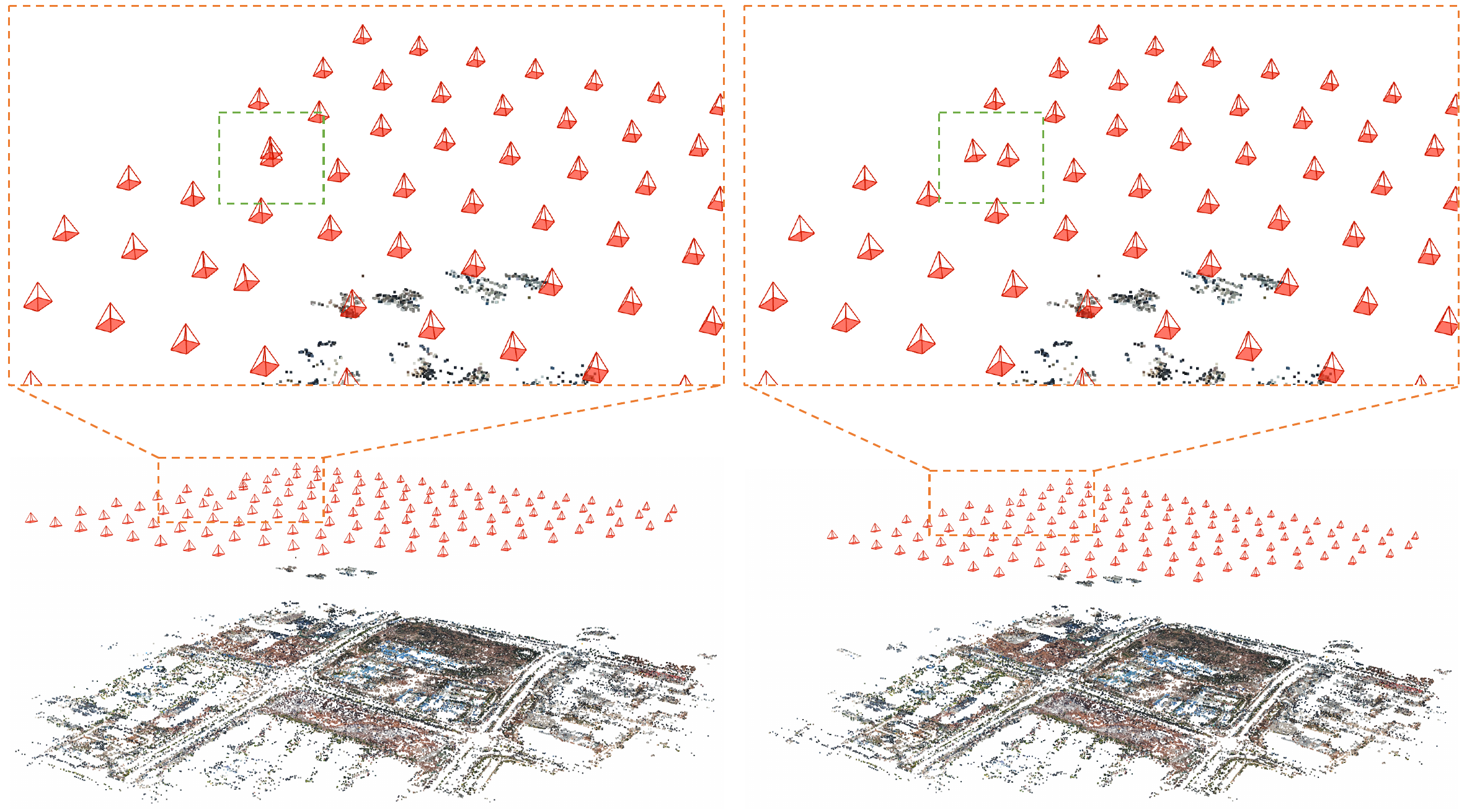}
\caption{BA results without outlier filtering (Left) and with outlier filtering (Right).}\label{fig:comp_woorba}
\end{figure}

(3) Upon removal of the global consistency-based outlier rejecting module, there is an increase in the reprojection error. As shown in Fig.\ref{fig:comp_woor}, the noise feature points generated by triangulation increases significantly. The outliers included in the scene can have a negative impact on subsequent BA and easily lead to a local optimum. For example, for scene 8\_127, the final result after BA is shown in Fig.\ref{fig:comp_woorba}. It can be seen that there is a camera (marked with a green dashed box) whose pose has been optimized erroneously due to the influence of outliers.

\subsection{Generalization experiments on different sizes of input images}
The proposed network does not require a fixed number of images as input. Therefore, to test the generalization of the network in scenarios with inconsistent camera numbers compared to the training sample, we apply the model trained on scene from 100-130 images to predict scenes with 30-50 cameras and 400-430 cameras. The predicted results are shown in Tab.\ref{table:general_or} and Tab.\ref{table:general_pp}.

\begin{table}[]
\caption{Results of outlier rejecting for 100-130 camera models in different numbers of image prediction tasks}\label{table:general_or}
\centering
\resizebox{0.48\textwidth}{!}{
\begin{tabular}{ccccccc}
\hline
Number of images        & Scene  & ↑Acc & ↑Pre & ↑Rec & ↑F1 & ↓Time/s \\ \hline
\multirow{6}{*}{30-50}   & 1\_45  & 0.952     & 0.968      & 0.973   & 0.971     & 0.720   \\
                         & 2\_42  & 0.964     & 0.976      & 0.981   & 0.979     & 0.713   \\
                         & 3\_50  & 0.950     & 0.961      & 0.976   & 0.969     & 0.749   \\
                         & 4\_34  & 0.920     & 0.953      & 0.947   & 0.950     & 0.698   \\
                         & 5\_35  & 0.942     & 0.970      & 0.955   & 0.962     & 0.699   \\
                         & Mean   & 0.946     & 0.966      & 0.967   & 0.966     & 0.716   \\ \hline
\multirow{6}{*}{400-430} & 1\_420 & 0.971     & 0.973      & 0.99    & 0.982     & 1.186   \\
                         & 2\_428 & 0.976     & 0.977      & 0.993   & 0.985     & 1.188   \\
                         & 3\_406 & 0.977     & 0.979      & 0.993   & 0.986     & 1.148   \\
                         & 4\_401 & 0.975     & 0.976      & 0.993   & 0.985     & 1.152   \\
                         & 5\_404 & 0.977     & 0.978      & 0.994   & 0.986     & 1.139   \\
                         & Mean   & 0.975     & 0.977      & 0.993   & 0.985     & 1.163   \\ \hline
\end{tabular}}
\end{table}

\begin{table*}[]
\caption{Pose prediction results of the 100-130 camera trained model in different numbers of camera prediction tasks}\label{table:general_pp}
\centering
\resizebox{0.8\textwidth}{!}{
\begin{tabular}{ccccccccc}
\hline
\multirow{2}{*}{Cam num} & \multirow{2}{*}{Scene} & \multirow{2}{*}{IPE/m} & \multicolumn{3}{c}{Results of AAT}                                & \multicolumn{3}{c}{Results after BA}                              \\ \cline{4-9} 
                                   &                        &                                 & ↓RPE/pix & ↓PE/m & ↓RE/° & ↓RPE/pix & ↓PE/m & ↓RE/° \\ \hline
\multirow{6}{*}{30-50}             & 1\_45                  & 5.309                           & 56.545                    & 4.719             & 2.055             & 0.466                     & 2.144             & 0.032             \\
                                   & 2\_42                  & 4.951                           & 48.010                    & 4.071             & 1.221             & 0.473                     & 2.291             & 0.029             \\
                                   & 3\_50                  & 5.245                           & 50.903                    & 3.983             & 1.701             & 0.475                     & 2.208             & 0.038             \\
                                   & 4\_34                  & 4.859                           & 52.996                    & 4.562             & 2.386             & 0.447                     & 1.172             & 0.035             \\
                                   & 5\_35                  & 5.448                           & 73.172                    & 4.393             & 2.586             & 0.415                     & 1.697             & 0.030             \\
                                   & Mean                   & 5.162                           & 56.325                    & 4.346             & 1.990             & 0.455                     & 1.902             & 0.033             \\
\hline
\multirow{6}{*}{400-430}           & 1\_420                 & 5.009                           & 63.770                    & 5.512             & 1.922             & 0.475                     & 3.132             & 0.046             \\
                                   & 2\_428                 & 5.111                           & 56.347                    & 4.928             & 1.884             & 0.482                     & 3.023             & 0.029             \\
                                   & 3\_406                 & 5.046                           & 54.624                    & 4.774             & 1.878             & 0.483                     & 3.164             & 0.028             \\
                                   & 4\_401                 & 4.919                           & 59.247                    & 4.905             & 1.872             & 0.486                     & 3.270             & 0.041             \\
                                   & 5\_404                 & 4.935                           & 59.912                    & 5.254             & 1.862             & 0.486                     & 3.148             & 0.034             \\
                                   & Mean                   & 5.004                           & 58.780                    & 5.075             & 1.884             & 0.482                     & 3.147             & 0.036             \\ \hline
\end{tabular}}
\end{table*}

\begin{figure*}[!t]
\centering
\includegraphics[width=0.9\linewidth]{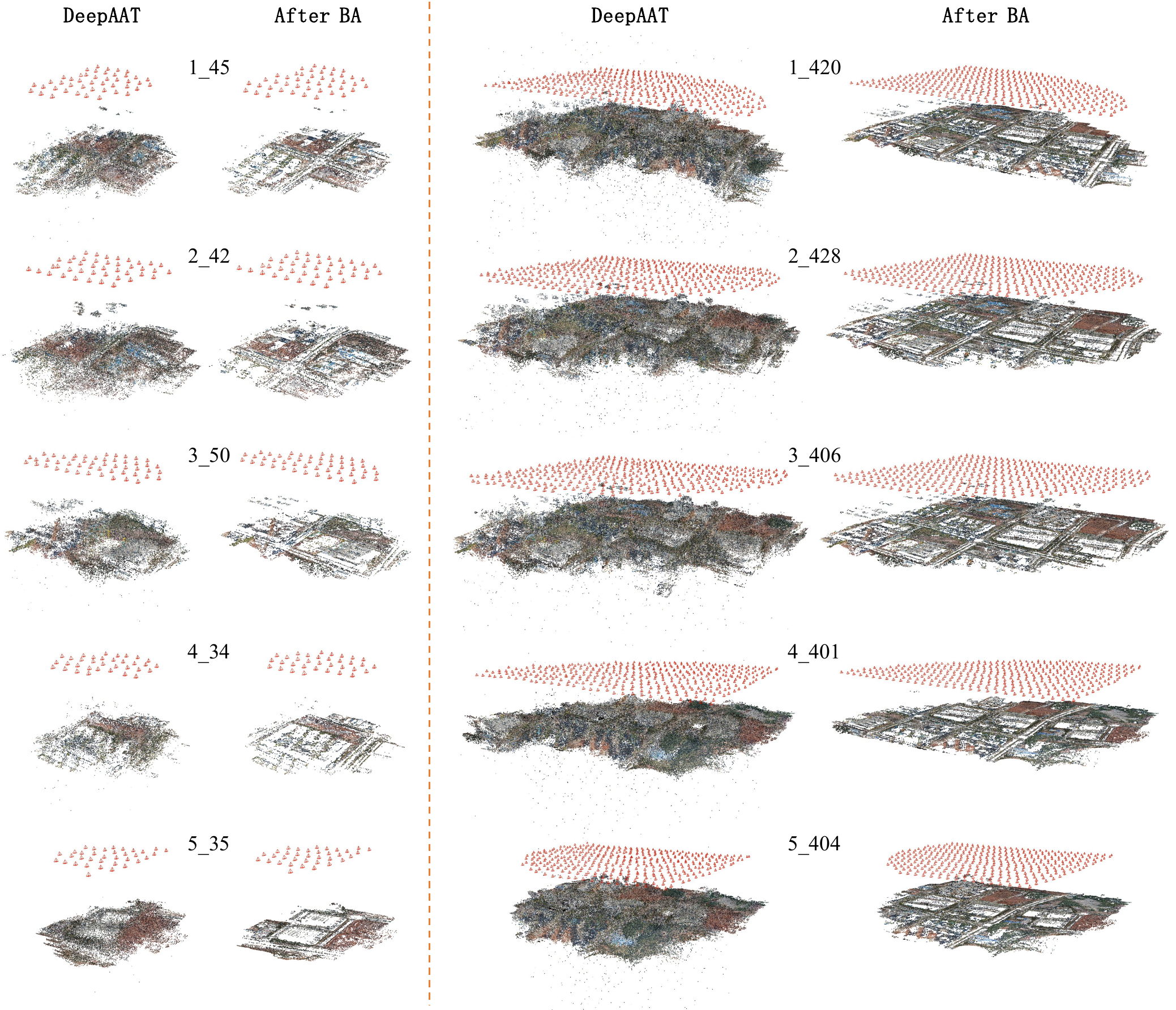}
\caption{Pose prediction results of the 100-130 camera trained model in different numbers of camera prediction tasks}\label{fig:generalization}
\end{figure*}

The prediction results demonstrate that the proposed DeepAAT is versatile, excelling not only in scenes with a similar number of images but also in scenarios with significantly more or fewer images. As indicated in Tab.\ref{table:general_or}, the network trained on scenes with 100-130 images shows a decline in accuracy, precision, recall, and F1 score when applied to 30-50 image scenes. Conversely, its performance metrics improve in 400-430 image scenes. This improvement could be attributed to the longer average track length of matching points in scenes with more cameras, increasing the likelihood of points being classified as inliers and reducing false negatives. This explains the high recall of 0.993 in scenes with 400-430 cameras. Regarding prediction time, the network demonstrates a notable efficiency: despite a rapid increase in the number of cameras, the time required for network prediction increases only marginally. This efficiency is a significant advantage in practical applications. In traditional AAT algorithms, both incremental and global, time consumption escalates quickly with an increasing number of scene images, a trend particularly pronounced in incremental SfM. Thus, our network's time-saving benefits become more pronounced with larger sets of scene images.

From Tab.\ref{table:general_pp}, we observe that the network, trained on scenes with 100-130 images, exhibits a slight increase in average reprojection error, position error, and rotation error while predicting scenes with 30-50 images and those with 400-430 images. However, in scenarios such as 1\_420 and 5\_404, the network's predicted position errors surpass the initial position errors. Nevertheless, following global BA, all scenes achieve accurate reconstruction results, as illustrated in Fig.\ref{fig:generalization}. This outcome not only underscores the network's effective scene initialization capabilities but also highlights its robust reconstruction prowess.

This experimental outcome indicates the superior performance of the proposed DeepAAT. It exhibits the capability to effectively handle scene prediction tasks several times larger than its training scope on smaller scenes. Typically, GPU memory consumption is higher during the training phase than in testing for most deep-learning tasks. Consequently, this characteristic significantly enhances the practicality of DeepAAT, making it a robust solution for large-scale applications.

\section{Conclusion} \label{section:conclusion}

AAT of UAV images has gained widespread adoption in 3D reconstruction, favored for its flexibility and cost-effectiveness. However, challenges persist: incremental AAT methods struggle with low reconstruction efficiency, global AAT methods grapple with subpar robustness and scene integrity, and deep learning-based algorithms often falter when processing a vast number of images. To overcome these challenges, we introduce DeepAAT, a novel approach designed to enhance the efficiency of UAV AAT while maintaining the accuracy and completeness of the reconstructed scenes. Our experiments demonstrate that DeepAAT's time efficiency outstrips incremental algorithms by hundreds of times and global algorithms by tens of times. In the near future, we will extend DeepAAT to the image set without GPS information.

\section{Acknowledgment}
	
This study was jointly supported by the National Natural Science Foundation Project (No. 42201477, No. 42130105).

\printcredits

\bibliographystyle{cas-model2-names}

\bibliography{cas-refs}

\end{document}